\pdfoutput=1

\documentclass[11pt]{article}

\usepackage[]{ACL2023}

\usepackage{hhline}
\usepackage{framed}
\usepackage{mathtools}

\usepackage{times}
\usepackage{latexsym}
\usepackage{multirow}
\usepackage{url}
\usepackage{amsmath}
\usepackage{array}
\usepackage{utfsym}
\usepackage{subfigure}
\usepackage{balance}
\usepackage[misc]{ifsym}
\usepackage{graphicx}
\usepackage{enumitem}
\usepackage{makecell}
\usepackage{booktabs}
\usepackage{amssymb}
\usepackage{color,xcolor}
\usepackage{cases}
\usepackage{colortbl} 
\usepackage{diagbox}
\definecolor{ggreen}{HTML}{d4e7cf}
\definecolor{bblue}{HTML}{00bfff}
\definecolor{ppink}{HTML}{ff69b4}
\newcommand{\modi}{\textcolor{black}}
\usepackage[T1]{fontenc}

\usepackage[utf8]{inputenc}

\usepackage{microtype}

\title{Zero-Shot Cross-Lingual Summarization via Large Language Models}

\author{Jiaan Wang\textsuperscript{1}\thanks{\ \ Equal Contribution. Work was done when Wang and Liang was interning at Pattern Recognition Center, WeChat AI, Tencent Inc, China.}\ \thanks{ \ \ Corresponding author.}, \ Yunlong Liang\textsuperscript{2}\footnotemark[1], \  Fandong Meng\textsuperscript{3}, \ Beiqi Zou\textsuperscript{4} \\
\bf {Zhixu Li\textsuperscript{5}, \ Jianfeng Qu\textsuperscript{1} and Jie Zhou\textsuperscript{3} } \\
\textsuperscript{1}Soochow University, Suzhou, China \quad \textsuperscript{2}Beijing Jiaotong University, Beijing, China\\
\textsuperscript{3}Pattern Recognition Center, WeChat AI, Tencent Inc, China \quad \textsuperscript{4}Princeton University, NJ, USA\\
\textsuperscript{5}Fudan Unversity, Shanghai, China \\
\texttt{jawang.nlp@gmail.com} \quad \texttt{yunlongliang@bjtu.edu.cn} \\
\texttt{fandongmeng@tencent.com} \quad \texttt{bzou@cs.princeton.edu}\\
}

\begin{document}
\maketitle
\begin{abstract}

Given a document in a source language, cross-lingual summarization (CLS) aims to generate a summary in a different target language.
Recently, the emergence of Large Language Models (LLMs), such as GPT-3.5, ChatGPT and GPT-4, has attracted wide attention from the computational linguistics community.
However, it is not yet known the performance of LLMs on CLS.
In this report, we empirically use various prompts to guide LLMs to perform zero-shot CLS from different paradigms (\emph{i.e.}, end-to-end and pipeline), and provide a preliminary evaluation on the generated summaries.
We find that ChatGPT and GPT-4 originally prefer to produce lengthy summaries with detailed information. These two LLMs can further balance informativeness and conciseness with the help of an interactive prompt, significantly improving their CLS performance.
Experimental results on three widely-used CLS datasets show that GPT-4 achieves state-of-the-art zero-shot CLS performance, and performs competitively compared with the fine-tuned mBART-50.

Moreover, we also find some multi-lingual and bilingual LLMs (\emph{i.e.}, BLOOMZ, ChatGLM-6B, Vicuna-13B and ChatYuan) have limited zero-shot CLS ability.
Due to the composite nature of CLS, which requires models to perform summarization and translation simultaneously, accomplishing this task in a zero-shot manner is even a challenge for LLMs.
\emph{Therefore, we sincerely hope and recommend future LLM research could use CLS as a testbed.}

\end{abstract}

\section{Introduction}

Cross-Lingual Summarization (CLS) aims to provide a target-language (\emph{e.g.}, Chinese) summary for a lengthy document in a different source language (\emph{e.g.}, English)~\cite{Leuski2003CrosslingualCE,wan-etal-2010-cross,yao-etal-2015-phrase,zhu-etal-2019-ncls,zhu-etal-2020-attend,ladhak-etal-2020-wikilingua,perez-beltrachini-lapata-2021-models,bai-etal-2021-cross,Liang2022AVH,feng-etal-2022-msamsum,Hasan2021CrossSumBE,Wang2022ClidSumAB,Wang2022ASO,liang2022summary,liu-etal-2022-assist,zheng2022long,aumiller-etal-2022-eur}. This task could help people quickly capture their interests from foreign documents.

\begin{figure}[t]
\centerline{\includegraphics[width=0.48\textwidth]{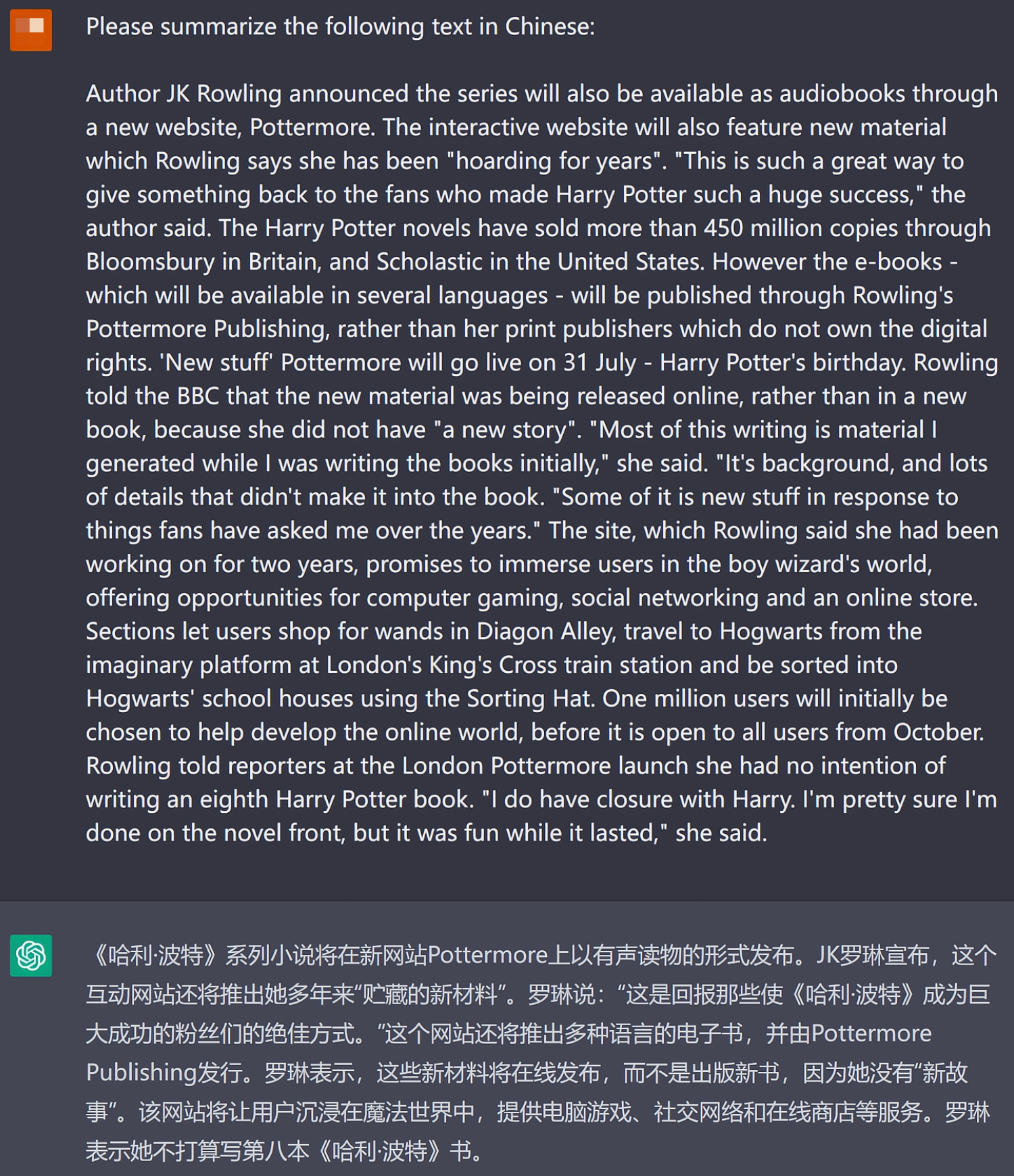}}
\caption{An example of zero-shot CLS via ChatGPT.}
\label{fig:example}
\vspace{-0.5cm}
\end{figure}

In recent years, a number of powerful multi-lingual pre-trained generative models have been proposed one after another, such as mBART~\cite{Liu2020MultilingualDP}, mBART-50~\cite{Tang2020MultilingualTW}, mT5~\cite{xue-etal-2021-mt5} and BLOOM~\cite{scao2022bloom}. The parameters in these models have gradually increased from million levels (\emph{e.g.}, 580M in mT5-base and 610M in mBART-Large) to billion levels (\emph{e.g.}, 3.7B in mT5-XL, 13B in mT5-XXL and 176B in BLOOM), facilitating various research topics (\emph{e.g.}, machine translation and CLS) in the multi-lingual world. Besides, large language models (LLMs) have been key to strong performance when transferring to new tasks by simply conditioning on a few input-label pairs (\emph{in-context learning})~\cite{dong2022survey,min-etal-2022-rethinking} or short sentences describing crucial reasoning steps (\emph{chain-of-thoughts})~\cite{fu2022complexity,zhang2022automatic}.

More recently, ChatGPT and GPT-4~\cite{OpenAI2023GPT4TR} have attracted great attention from both the research communities and industries. Similar to InstructGPT~\cite{ouyang2022training}, ChatGPT is created by fine-tuning a GPT-3.5 series model via reinforcement learning from human feedback (RLHF)~\cite{christiano2017deep}.
GPT-4, as a multi-modal LLM that can accept image and text inputs and produce text outputs, exhibits human-level performance on various benchmark datasets~\cite{OpenAI2023GPT4TR}.
With the emergence of ChatGPT and GPT-4, there is growing interest in leveraging LLMs for various NLP tasks~\cite{qin2023chatgpt,jiao2023chatgpt,bang2023multitask,Yang2023ExploringTL,zhong2023can,wang2023chatgpt,bubeck2023sparks,tan2023evaluation,peng2023towards,liu2023gpteval,yong2023prompting}.
Nevertheless, the exploration of LLMs on CLS is still lacking.

In this report, we present a preliminary evaluation of LLMs' zero-shot CLS performance, including GPT-3.5, ChatGPT, GPT-4, BLOOMZ, ChatGLM-6B, Vicuna-13B and ChatYuan.
In detail, we design various prompts to guide LLMs to perform CLS in an end-to-end manner with or without chain-of-thoughts (CoT).
Figure~\ref{fig:example} gives an example of prompting ChatGPT to perform zero-shot CLS.
To further exploit the interaction capability of conversational LLMs (\emph{e.g.}, ChatGPT and GPT-4), we leverage an interactive prompt to let them produce more concise summaries.
Moreover, to provide a deeper analysis of LLMs' zero-shot CLS performance, we compare them with fine-tuned mBART-50~\cite{Tang2020MultilingualTW} which has shown its superiority in many previous CLS works~\cite{Wang2022ClidSumAB,feng-etal-2022-msamsum,perez-beltrachini-lapata-2021-models}.
Experimental results on three CLS datasets, covering three domains (news, how-to guide and dialogue) and two cross-lingual directions (En$\Rightarrow$Zh and En$\Rightarrow$De)\footnote{Since a CLS dataset might contain multiple source and target languages, we use ``X$\Rightarrow$Y'' to indicate the source language and target language are X and Y, respectively. En: English; Zh: Chinese; De: German.}, show that GPT-4 achieves the best zero-shot performance but is still worse than the fine-tuned mBART-50 model in terms of ROUGE scores and BERTScore. We also conduct case studies to show that ChatGPT and GPT-4 could absorb the core idea of the given source-language documents and generate fluent and concise target-language summaries.

In addition, we find that the current open-source LLMs (\emph{i.e.}, BLOOMZ, ChatGLM-6B, Vicuna-13B and ChatYuan) achieve limited zero-shot CLS performance, which is significantly worse than that of GPT-4.
This leads us to conclude that the composite end-to-end CLS prompts are difficult for them to follow, and there is still a challenge for LLMs to perform zero-shot CLS in an end-to-end manner which requires simultaneously carrying out translation and summarization.
Based on the finding, we suggest that future multi-lingual or bilingual LLM research uses CLS as a testbed to evaluate LLMs' capabilities to follow composite instructions as well as combine their different abilities.

Our main contributions are concluded as follows:
\begin{itemize}[leftmargin=*,topsep=0pt]
\setlength{\itemsep}{0pt}
\setlength{\parsep}{0pt}
\setlength{\parskip}{0pt}
\item To the best of our knowledge, we are the first to explore the zero-shot CLS performance of LLMs. To achieve that, we design various prompts to guide LLMs to perform CLS in an end-to-end manner with or without chain-of-thoughts.
\item Experimental results on three widely-used CLS benchmark datasets, covering various domains and languages, show several LLMs (especially ChatGPT and GPT-4) achieve competitive results compared with the strong fine-tuned baseline.
\item We also find the current open-source LLMs generally achieve limited zero-shot CLS performance, making us think CLS could be used as a testbed for future LLM research due to its challenges.
\end{itemize}

\begin{figure*}[t]
\centerline{\includegraphics[width=0.98\textwidth]{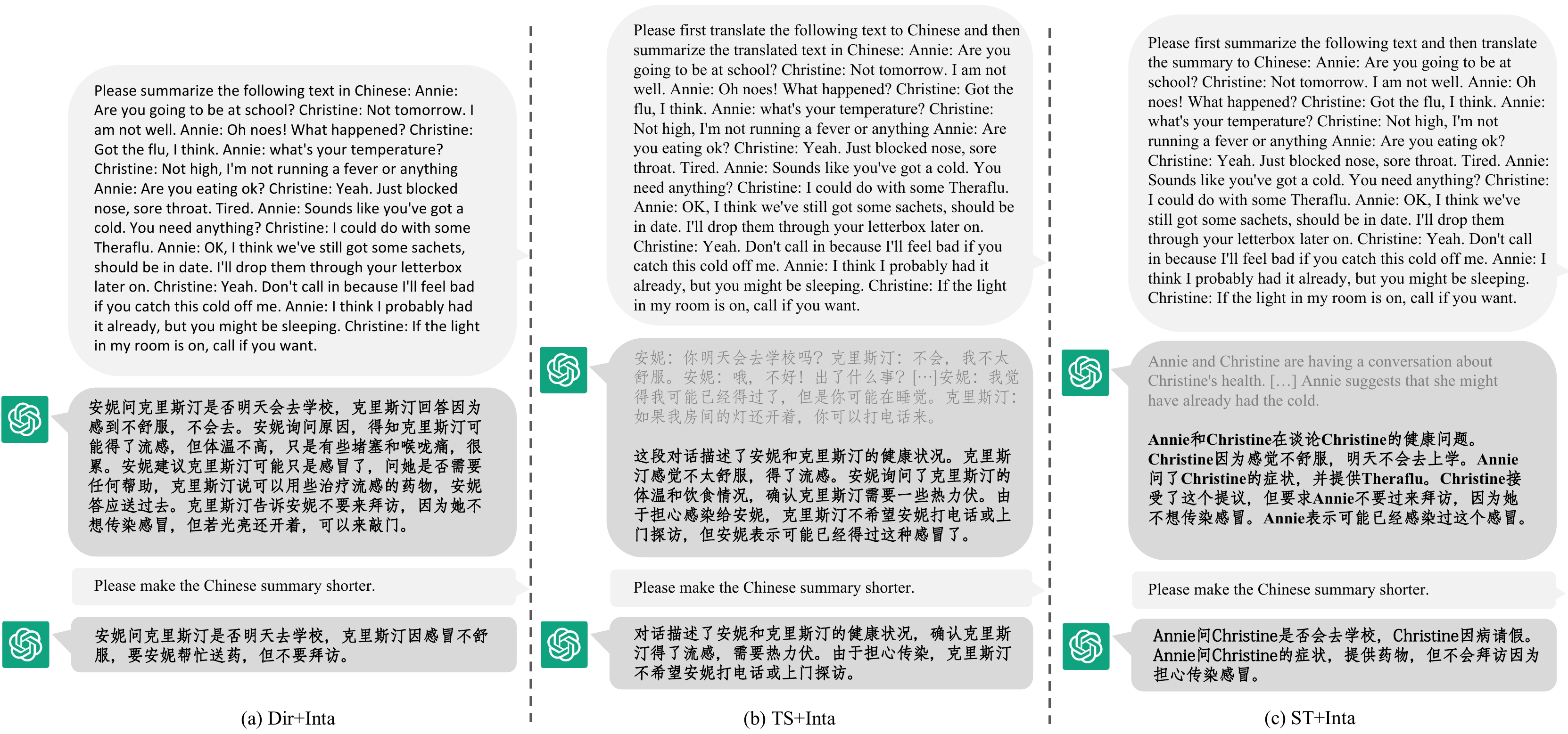}}
\caption{An illustration of all prompts used to guide LLMs to perform zero CLS (take ChatGPT as an example).}
\label{fig:prompts}
\end{figure*} 

\section{Methodology}
\subsection{Cross-Lingual Summarization Prompts}
\label{sec: evaluation_method}
We heuristically design 3 prompts to guide LLMs to perform zero-shot CLS in an end-to-end manner, which is shown as follows with an example from an English document to a Chinese summary:

\vspace{0.5ex}
\begin{itemize}[leftmargin=*,topsep=0pt]
\item The direct (\textbf{Dir}) prompt guides LLMs straightforwardly output the corresponding target-language summary without chain-of-thoughts (CoT):

\begin{quote}
\texttt{Please summarize the following text in Chinese: [English~Doc]}
\end{quote}
where \texttt{[English Doc]} indicates a given English document.

\item The translate-then-summarize (\textbf{TS}) CoT prompt makes LLMs first translate the given document from the source language to the target language, and then summarize the translated document to perform CLS:

\begin{quote}
\texttt{Please first translate the following text to Chinese and then summarize the translated text in Chinese: [English~Doc]}
\end{quote}

\item The summarize-then-translate (\textbf{ST}) CoT prompt lets LLMs first summarize the given document and then translate the output summary to the target language:

\begin{quote}
\texttt{Please first summarize the following text and then translate the summary to Chinese: [English~Doc]}
\end{quote}
\end{itemize}

Note that though the TS and ST CoT prompts guide LLMs to perform CLS step by step, the behaviors are end-to-end since the target-language summaries are generated within a single turn.

To further exploit the potentiality of conversational LLMs, inspired by~\citet{bang2023multitask}, after prompting with Dir, TS or ST prompt, we adopt an interactive (\textbf{Inta}) prompt to make the preliminarily generated summary more concise:

\begin{quote}
\texttt{Please make the Chinese summary shorter.}
\end{quote}
and the whole process is denoted as ``\textbf{Dir+Inta}'', ``\textbf{TS+Inta}'' or ``\textbf{ST+Inta}''.

\subsection{Large Language Models}

We explore the CLS ability of the following LLMs:

\begin{table*}[t]
\centering
\resizebox{0.98\textwidth}{!}
{
\begin{tabular}{lcccccc}
\toprule[1pt]
\multicolumn{1}{c}{Dataset} & Src Lang.                & Trg Lang. & Domain                        & Example            & Doc. Length & Sum. Length \\ \midrule[1pt]
CrossSum                    & English                  & Chinese   & News                          & 3981 / 497 / 50 out of 497     & 814.2       & 35.6        \\ \midrule
\multirow{2}{*}{WikiLingua} & \multirow{2}{*}{English} & Chinese   & \multirow{2}{*}{How-to guide} & 13211 / 1886 / 50 out of 3775  & 538.6       & 53.2        \\
                            &                          & German    &                               & 40839 / 5833 / 50 out of 11669 & 526.1       & 63.4        \\ \midrule
\multirow{2}{*}{XSAMSum}    & \multirow{2}{*}{English} & Chinese   & \multirow{2}{*}{Dialogue}     & 14732 / 818 / 50 out of 819        & 140.1       & 27.6        \\
                            &                          & German    &                               & 14732 / 818 / 50 out of 819        & 140.1       & 31.7        \\ \bottomrule[1pt]
\end{tabular}
}
\caption{Statistics of CLS datasets used in experiments. ``\emph{Src Lang.}'' and ``\emph{Trg Lang}'' denote the source and the target languages. ``\emph{Doc. Length}'' and ``\emph{Sum. Length}'' show the average length of source documents and target summaries (token level). ``\emph{Example}'' lists the number of samples in each dataset w.r.t training, validation and test sets.}
\label{table:statistics}
\end{table*}

\begin{itemize}[leftmargin=*,topsep=0pt]
\setlength{\parsep}{0pt}
\setlength{\parskip}{0pt}
\item \textbf{Davinci-003} is the most advanced GPT-3.5 model with 175B parameters. We evaluate its performance by requesting the official API provided by OpenAI with default settings.\footnote{\url{https://platform.openai.com/docs/models/gpt-3-5}}
\item \textbf{ChatGPT} is created by fine-tuning a GPT-3.5 series model via reinforcement learning from human feedback (RLHF)~\cite{christiano2017deep}. We conduct experiments on the ChatGPT platform\footnote{\url{https://chat.openai.com/}} between February 17 to February 19, 2023.
\item \textbf{GPT-4}, as a multi-modal LLM that can accept image and text inputs and produce text outputs, exhibits human-level performance on various benchmark datasets~\cite{OpenAI2023GPT4TR}. We assess GPT-4 on the ChatGPT platform between March 15 to March 19, 2023.
\item \textbf{BLOOMZ}~\cite{muennighoff2022crosslingual} is an open-source multi-lingual LLM with 176B parameters. The model supports 59 languages, and is created by fine-tuning BLOOM~\cite{scao2022bloom} on an instruction corpus (\emph{i.e.}, XP3).
\item \textbf{ChatGLM-6B}\footnote{\url{https://github.com/THUDM/ChatGLM-6B}} is an open-source bilingual (\emph{i.e.}, Chinese and English) language model based on General Language Model (GLM) framework~\cite{du2022glm}. The model suffers from both instruction tuning and RLHF. 
\item \textbf{Vicuna-13B}\footnote{\url{https://vicuna.lmsys.org/}} is an open-source LLM created by fine-tuning LLaMA~\cite{touvron2023llama} on user-shared conversations collected from ChatGPT. We evaluate the model via its demo platform\footnote{\url{https://chat.lmsys.org/}} between March 31 to April 2, 2023.
\item \textbf{ChatYuan}\footnote{\url{https://github.com/clue-ai/ChatYuan}} is an open-source bilingual (\emph{i.e.}, Chinese and English) LLM with 7.7B parameters. The training process of this model includes instruction tuning and RLHF.
\end{itemize}

Among the above LLMs, ChatGPT, GPT-4, ChatGLM-6B, Vicuna-13B and ChatYuan are conversational LLMs while Davinci-003 and BLOOMZ are not. When evaluating their zero-shot CLS performance, we only equip conversational LLMs with the interactive prompt.

\begin{table*}[t]
\centering
\resizebox{0.98\textwidth}{!}
{
\begin{tabular}{lcccccccccccccccccccc}
\toprule[1pt]
\multicolumn{1}{c}{\multirow{2}{*}{Model}} & \multicolumn{4}{c}{CrossSum (En$\Rightarrow$Zh)}                         & \multicolumn{4}{c}{WikiLingua (En$\Rightarrow$Zh)}                       & \multicolumn{4}{c}{WikiLingua (En$\Rightarrow$De)}                       & \multicolumn{4}{c}{XSAMSum (En$\Rightarrow$Zh)}                           & \multicolumn{4}{c}{XSAMSum (En$\Rightarrow$De)}                           \\
\cmidrule(r){2-5}\cmidrule(r){6-9}\cmidrule(r){10-13}\cmidrule(r){14-17}\cmidrule(r){18-21} \multicolumn{1}{c}{}                     & R-1           & R-2          & R-L           & B-S           & R-1           & R-2          & R-L           & B-S           & R-1           & R-2          & R-L           & B-S           & R-1           & R-2           & R-L           & B-S           & R-1           & R-2           & R-L           & B-S           \\ \midrule[1pt]
\cellcolor{ppink!20}{mBART-50}                               & 26.1          & 7.4          & 22.1          & 65.4          & 32.1          & 10.4         & 26.8          & 68.5          & 26.8          & 7.7          & 20.5          & 62.5          & 40.6          & 14.4          & 33.9          & 74.5          & 42.4          & 18.9          & 35.4          & 73.7          \\ \midrule[1pt]
\cellcolor{bblue!5}{ChatYuan-7.7B (Dir)}    &  0.3  & 0.0  & 0.3     &  51.7     &   4.1    &  1.2   &  2.6   &  54.4 &  -  &  -    &   -    &     -   &   0.8    & 0.3  & 0.7  &    48.5   &  -  &  -    &   -    &     -  \\
\cellcolor{bblue!5}{ChatYuan-7.7B (Dir+Inta)}    & 0.2   & 0.0  &  0.2    &     52.0  &   4.7    & 1.6    &  3.4   & 51.6  & -   &   -   &    -   &    -  &   0.3   &  0.1     & 0.3  &  47.0 &  -  &  -    &   -    &     -   \\
\cellcolor{bblue!5}{ChatYuan-7.7B (TS)}    & 0.4   & 0.0  & 0.4     &   46.6    &    8.2   &  2.7   & 5.4    & 56.0  &  -  &  -    &   -    &    -  &   11.3   &   4.2    &  8.7 & 49.5 &  -  &  -    &   -    &     -     \\ 
\cellcolor{bblue!5}{ChatYuan-7.7B (TS+Inta)}    & 2.0   & 0.5  &  1.4    &   46.9    &   6.9    &  2.1   &  4.3   & 53.4  & -   &   -   &     -  &    -  &    9.5  &  3.2     & 6.9  & 52.3 &  -  &  -    &   -    &     -     \\ 
\cellcolor{bblue!5}{ChatYuan-7.7B (ST) }   &  0.5  & 0.0  &  0.4    &   49.6    &    6.9   &  2.1   &  4.2   & 56.1  &   - &    -  &   -    &  -    &   7.5   &  2.5     & 5.5  &  49.6&  -  &  -    &   -    &     -   \\ 
\cellcolor{bblue!5}{ChatYuan-7.7B (ST+Inta)}    &  1.2  &  0.4 &  0.9    &   49.7    &  7.3     & 2.3    &  4.6   & 55.5  &   - &   -   &   -    &   -   &   6.0   &  2.0     & 3.9  & 48.5 &  -  &  -    &   -    &     -    \\ 
\cellcolor{bblue!5}{ChatGLM-6B (Dir)}    &  5.7  & 2.3  &  2.4    &   53.9    &   14.5    &  5.3   &  9.9   & 59.5  &  -  &    -  &   -    &    -  &  20.4    &   9.1    &  15.3 & 58.8 &  -  &  -    &   -    &     -    \\
\cellcolor{bblue!5}{ChatGLM-6B (Dir+Inta) }   & 7.9   & 2.4  &  5.3    &  55.6     &   14.6    &  5.1   & 9.5    & 59.1  & -   &  -    &    -   &   -   &  18.0    &  8.0     & 14.0  & 59.5  &  -  &  -    &   -    &     -    \\
\cellcolor{bblue!5}{ChatGLM-6B (TS)}    &  8.4  & 2.9  & 4.8     &  54.1     &  14.6     &  5.3   & 9.8    & 59.7  &   - &   -   &  -     &    -  &   21.5   &  9.6     & 16.6  &   57.9   &  -  &  -    &   -    &     -     \\ 
\cellcolor{bblue!5}{ChatGLM-6B (TS+Inta)}    &  9.6  & 3.0  &  6.1    &   55.2    &    14.9   &  5.1   &  9.4   & 59.1  &  -  &   -   &  -     &   -   &  18.7    &   8.1    & 15.0  & 58.6  &  -  &  -    &   -    &     -      \\ 
\cellcolor{bblue!5}{ChatGLM-6B (ST)}   &  5.8  &  1.8 &   3.6   &   53.2    &    15.6   &  5.5   &  10.2   & 59.9  &  -  &    -  &  -     &     - &   19.8   &  8.3     &  14.7 & 58.1  &  -  &  -    &   -    &     -     \\ 
\cellcolor{bblue!5}{ChatGLM-6B (ST+Inta) }   &  2.2  & 0.6  &   1.7   &  53.8     &    9.8   & 3.3    &  6.1   & 57.0  &  -  &   -   &   -    &   -   &    12.7  &  5.1     & 9.9  & 56.8  &  -  &  -    &   -    &     -    \\ 
\cellcolor{bblue!5}{Vicuna-13B (Dir)}  &  -  &   -   &   -    &   - &  -  &   -   &   -    &   -   &  -  &    -  &   -    &    -  &   19.5   &  7.2   &  14.5    &  60.1  &   22.5   &  4.9   &  17.6    &  58.5   \\
\cellcolor{bblue!5}{Vicuna-13B (Dir+Inta) }  &  -  &   -   &   -    &   - &  -  &   -   &   -    &   -  & -   &  -    &    -   &   -   &   24.1   &  9.7   &  18.9    & 63.0  &  28.7    &  7.8   &  22.0    &  60.5  \\
\cellcolor{bblue!5}{Vicuna-13B (TS)}   &  -  &   -   &   -    &   - &  -  &   -   &   -    &   - &   - &   -   &  -     &    - &   18.3   &   7.1  &   14.6   & 61.6  &  25.0    &  5.9   &   18.2   &  59.4  \\ 
\cellcolor{bblue!5}{Vicuna-13B (TS+Inta)}    &  -  &   -   &   -    &   - &  -  &   -   &   -    &   -  &  -  &   -   &  -     &   -  &   22.0   &  7.9   &   17.4   &  64.3 &   31.7   &   8.9  &   24.2   &   61.2   \\ 
\cellcolor{bblue!5}{Vicuna-13B (ST)}  &  -  &   -   &   -    &   - &  -  &   -   &   -    &   -   &  -  &    -  &  -     &     - &   17.5   &  6.1   &    13.6  &  59.6 &   27.3   &  6.8   &    20.4  &  59.3    \\ 
\cellcolor{bblue!5}{Vicuna-13B (ST+Inta) }   &  -  &   -   &   -    &   - &  -  &   -   &   -    &   -   &  -  &   -   &   -    &   - &    19.8  &  7.4   &    15.4  &  62.2 &  31.6    &  9.4   &  24.1    & 61.9   \\ 
\cellcolor{bblue!5}{BLOOMZ-176B (Dir) }   &  0.7  & 0.1  &  0.7    &    29.2   &   0.3    &  0.0   &  0.2   &  8.9 &  0.0  &   0.0   &   0.0    &    3.3     &   21.4    &  11.2 & 17.8  &  65.3      &     13.0   &    1.2     &   11.9   &  56.2  \\
\cellcolor{bblue!5}{BLOOMZ-176B (TS) }   &  2.1  & 1.3  &  1.6    &   21.5    &    0.4   &  0.0   &  0.3   & 5.6  &  0.0  &   0.0   &   0.0    &  5.0    &   30.4   &   \cellcolor{ggreen!60}{15.0}    & 25.2  &  64.8 &   12.5     &    0.7    &  11.4       &   54.4     \\ 
\cellcolor{bblue!5}{BLOOMZ-176B (ST)}    &  3.0  &  1.2 &  2.4    &  33.8     &    0.3   &  0.0   &  0.2   & 9.0  &  0.0  &  0.0    &   0.0    &  3.3    &  28.1    &  13.4     &  23.4 & 66.3  &  13.8      &   1.3     &  12.8       &   54.8   \\ \midrule[1pt]
\cellcolor{bblue!10}{Davinci-003 (Dir)}                           & 18.7          & 3.6          & 14.7          & 60.2          & 23.6          & 3.8          & 17.8          & 60.9          & 18.8          & 2.6          & 12.2          & 60.7          & 24.4          & 8.0           & 20.7          & 63.4          & 35.5          & 12.4          & 27.3          & 62.4          \\
\cellcolor{bblue!10}{Davinci-003 (TS)}        &  22.9  & \cellcolor{ggreen!60}{8.9}  &   13.5   &    59.6   &   23.7    &  8.2   &  15.1   & 61.0  & 16.9   &  2.0    &  10.9     &  59.2    &  33.3    &   \cellcolor{ggreen!60}{17.1}    &  26.6 & 64.7  &   34.7     &    11.5    &   26.1      &   62.0    \\
\cellcolor{bblue!10}{Davinci-003 (ST)}          &  \cellcolor{ggreen!60}{\textbf{26.2}}  & \cellcolor{ggreen!60}{\textbf{9.3}}  &  \textbf{16.9}    &    61.3   &       24.2 &   8.4  &   15.9  & 61.2  & 19.8   &  2.8    &   13.1    &  60.4    &   34.1   &   \cellcolor{ggreen!60}{\textbf{18.2}}    & 26.4  &  68.1 &    35.7    &   11.7     &    26.9     &    63.0   \\
\cellcolor{bblue!10}{ChatGPT (Dir) }                             & 14.2          & 3.3          & 10.3          & 60.3          & 20.9          & 5.6          & 15.5          & 62.7          & 16.9          & 2.1          & 10.7          & 60.1          & 21.3          & 5.5           & 17.1          & 63.5          & 32.0          & 10.3          & 24.5          & 61.4          \\
\cellcolor{bblue!10}{ChatGPT (Dir+Inta) }               & 22.1          & 3.8          & 15.6          & 61.8          & 28.4          & 6.5          & 22.1          & 64.5          & 22.4 & 2.8          & 14.7          & 61.3 & 27.2          & 6.9           & 22.9          & 67.5          & 39.6 & \textbf{16.0} & 31.4 & \textbf{64.3} \\
\cellcolor{bblue!10}{ChatGPT (TS) }                        & 15.8          & 3.3          & 11.9          & 60.9          & 24.8          & 5.4          & 19.1          & 62.9          & 19.4          & 2.4          & 12.6          & 60.0          & 26.0          & 7.3           & 21.2          & 66.4          & 33.2          & 9.6           & 25.3          & 61.1          \\
\cellcolor{bblue!10}{ChatGPT   (TS+Inta)}         & 22.6 & 4.1 & \textbf{16.9} & 62.7 & 26.1          & 5.3          & 19.7          & 63.7          & 21.6          & 2.4          & 15.1          & 60.8          & 27.4          & 6.7           & 22.4          & 67.1          & 39.4          & 13.5          & 29.4          & 63.3          \\
\cellcolor{bblue!10}{ChatGPT (ST) }                        & 16.5          & 3.8          & 12.0          & 60.8          & 27.2          & 7.3          & 20.3          & 64.3          & 21.3          & 3.5 & 14.4          & 60.9          & 26.8          & 7.7           & 21.3          & 66.7          & 31.7          & 8.8           & 23.5          & 60.8          \\
\cellcolor{bblue!10}{ChatGPT   (ST+Inta) }        & 21.6          & 3.5          & 15.5          & 61.7          & 30.1 & 8.1 & \textbf{22.4} & 64.9 & 21.4          & 3.1          & 15.4 & 60.6          & 31.4 & 11.5 & 28.1 & 70.1 & 35.9          & 13.2          & 29.0          & 62.8          \\ 
\cellcolor{bblue!10}{GPT-4 (Dir)}     & 13.7 & 3.7   & 10.1    &  59.7  & 23.1  & 9.1  & 15.5 & 63.5      & 20.4  & 3.3 & 13.8  & 62.2       & 24.5 & 7.1  & 19.5  & 66.1   &  34.7 & 13.4  & 25.3 & 61.7        \\
\cellcolor{bblue!10}{GPT-4 (Dir+Inta) }    & 20.3 & 4.4   & 14.1  & 61.9  & 30.4  & \cellcolor{ggreen!60}{11.7}  & 20.9 & \textbf{65.7}             & \textbf{24.8} & 3.9  & 17.0  & \cellcolor{ggreen!60}{\textbf{63.5}}  & 31.3  & 7.3 & 26.5  & 70.7    & \textbf{40.5}  & 13.4  & 30.8 & 64.2        \\
\cellcolor{bblue!10}{GPT-4 (TS) }    & 19.4  &  3.6  & 14.3   & 60.9  & 28.5  & \cellcolor{ggreen!60}{11.4}  & 18.2 & 64.2      & 23.1  & 3.8 & 16.3  & \cellcolor{ggreen!60}{62.7}       & \textbf{34.7} & 12.5  & \textbf{28.5}  & 71.0     & 38.9  & 11.9  & 29.0 & 63.3        \\
\cellcolor{bblue!10}{GPT-4 (TS+Inta) }    & 22.7 & 4.3   & 16.1  & 62.2  & 29.2  & \cellcolor{ggreen!60}{\textbf{12.6}}  & 20.3 & 64.9      & 23.6  & 3.9 & 17.3  & \cellcolor{ggreen!60}{62.9}       & 30.8 & 6.5  & 25.6  & 70.9     & 39.1  & 13.5  & \textbf{32.6} & 64.1        \\
\cellcolor{bblue!10}{GPT-4 (ST) }    & 19.0 & 4.3   & 14.1  & 61.7  & 30.2  & \cellcolor{ggreen!60}{12.2}  & 19.5 & 64.2      & 23.4  & 3.8 & 16.4  & \cellcolor{ggreen!60}{63.0}       & 32.1 & 10.7  & 26.4  & 70.7       & 38.6  & 12.3  & 29.5 & 63.2        \\
\cellcolor{bblue!10}{GPT-4 (ST+Inta) }    & 22.6 &  4.9  & 16.8  & \textbf{63.1}  & \textbf{30.5}  & \cellcolor{ggreen!60}{11.9}  & 21.3 & 65.2      & 23.1  & \textbf{4.2} & \textbf{17.4}  & \cellcolor{ggreen!60}{62.7}       & 29.2 & 8.2  & 25.4  & \textbf{71.4}     & 39.0  & 11.5  & 31.2 & 63.7        \\ \bottomrule[1pt]

\end{tabular}
}
\caption{Experimental results on CrossSum, WikiLingua and XSAMSum. \colorbox{ppink!20}{Pink} denotes the fine-tuned baseline. \colorbox{bblue!5}{Light blue} and \colorbox{bblue!10}{blue} denote the zero-shot performance of open-source and non-open-source LLMs, respectively. \colorbox{ggreen!60}{Green} indicates the zero-shot result is better than that of the fine-tuned baseline. ``-'' denotes the model cannot be evaluated in the corresponding dataset.}
\label{table:experiments}
\end{table*}

\section{Experiments}

\subsection{Experimental Setup}
\noindent \textbf{Datasets.} We evaluate LLMs on the following three CLS datasets: CrossSum (En$\Rightarrow$Zh)~\cite{Hasan2021CrossSumBE}, WikiLingua (En$\Rightarrow$Zh/De)~\cite{ladhak-etal-2020-wikilingua} and XSAMSum (En$\Rightarrow$Zh/De)~\cite{Wang2022ClidSumAB}. CrossSum is collected from BBC news website, it contains 3,981 English news reports paired with Chinese summaries. WikiLingua involves 18,887 English how-to guides paired with Chinese summaries, and 58,375 English how-to guides paired with German summaries.
Note that both CrossSum and WikiLingua also provide CLS samples in other cross-lingual directions, and we only utilize En$\Rightarrow$Zh or (and) En$\Rightarrow$De samples in this work.
XSAMSum contains 16,369 English dialogues paired with both Chinese and German summaries. The detailed statistics of these datasets are listed in Table~\ref{table:statistics}.
Since ChatGPT, GPT-4 and Vicuna-13B can only be interacted with manually when we conduct experiments, evaluating their performance is time-consuming.
Thus, we randomly sample 50 documents from the test set of each CLS dataset for evaluation. 

\vspace{0.5ex}
\noindent \textbf{Metrics.} We adopt ROUGE-1/2/L (R-1/2/L)~\cite{Lin2004ROUGEAP} and BERTScore (B-S)~\cite{Zhang2020BERTScoreET} in our experiments. The ROUGE scores measure the lexical overlap between the generated summaries and corresponding references based on the unigram,  bigram and longest common subsequence, while the BERTScore measures the semantic similarity. For ROUGE scores, we use \textit{multi-lingual rouge}\footnote{\url{https://github.com/csebuetnlp/xl-sum/tree/master/multilingual_rouge_scoring}} toolkit. For BERTScore, we use \textit{bert-score}\footnote{\url{https://github.com/Tiiiger/bert_score}} toolkit, and the score is calculated based on \textit{bert-base-multilingual-cased}\footnote{\url{https://huggingface.co/bert-base-multilingual-cased}} model.

\vspace{0.5ex}
\noindent \textbf{Baselines.} We also compare zero-shot LLMs with fine-tuned mBART-50~\cite{Tang2020MultilingualTW} to provide a deeper analysis. mBART-50 is a multi-lingual version of BART~\cite{lewis-etal-2020-bart} with the vanilla transformer encoder-decoder architecture~\cite{vaswani2017attention}. This model has been pre-trained on large-scale multi-lingual unlabeled corpora with BART-like denoising objectives.

\subsection{Implementation Details}

For ChatGPT, GPT-4 and Vicuna-13B, we manually evaluate their results via the corresponding platform and demo websites.
Among them, the demo website of Vicuna-13B cannot support the long input sequences, and it will automatically truncate the long sequences, thus we only evaluate Vicuna-13B on XSAMSum (En$\Rightarrow$Zh/De).
For Davinci-003, we use the official API with default settings.

For BLOOMZ, ChatGLM-6B and ChatYuan, we download the corresponding checkpoints and evaluate their performances following the officially released codes.
The 176B BLOOMZ makes use of 5*80G GPUs to load with FP16 precision.
We use a sampling decoding strategy and set the temperature to 0.7.
We only evaluate ChatGLM-6B and ChatYuan on En$\Rightarrow$Zh cross-lingual direction due to their bilingualism (\emph{i.e.}, Chinese and English).

For mBART-50 baseline, inspired by~\citet{feng-etal-2022-msamsum} and \citet{Wang2022ClidSumAB}, we employ \texttt{mBART-large-50-many-to-many-mmt} model\footnote{\url{https://huggingface.co/facebook/mbart-large-50-many-to-many-mmt}} via the implementation of Huggingface Transformers~\cite{wolf-etal-2020-transformers}. This model is fine-tuned with 4 batch size, 5e-6 learning rate and 10 epochs on each CLS dataset.

\begin{figure*}[t]
\centerline{\includegraphics[width=0.98\textwidth]{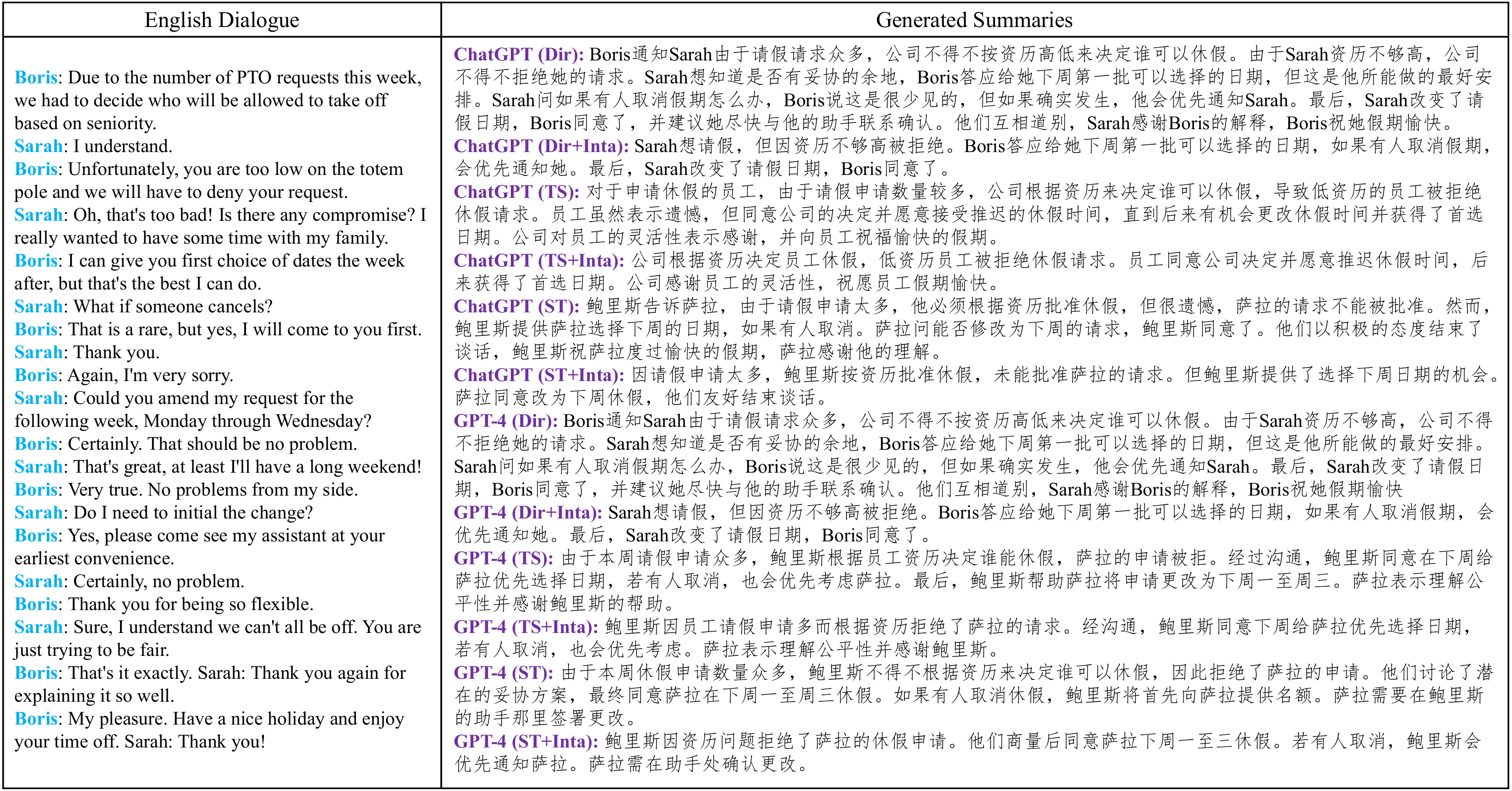}}
\caption{Example dialogue document in XSAMSum and summaries generated by ChatGPT and GPT-4.}
\label{fig:cases}
\end{figure*} 

\subsection{Main Results}

Table~\ref{table:experiments} lists the experimental results. As we can see, Davinci-003, ChatGPT and GPT-4 achieve competitive results with the fine-tuned mBART-50.

\vspace{0.5ex}
\noindent \textbf{The Effect of Each CLS Prompt.}
Among three end-to-end prompts (\emph{i.e.}, Dir, ST and TS), the CoT prompts lead to better performance than the direct prompt, indicating the effectiveness of CoT.
It also indicates that it is still challenging for a single model to directly perform CLS without giving any crucial or helpful instructions.

\vspace{0.5ex}
\noindent \textbf{The Effect of Interactive Prompt.}
Further, with the help of the interactive prompt, the performance of ChatGPT and GPT-4 significantly improve and even outperform mBART-50 in several automatic metrics. As shown in Table~\ref{table:output_lens}, more concise summaries can be generated after inputting the interactive prompt, \emph{e.g.}, 183.7 tokens generated by ChatGPT (Dir) on CrossSum, while the counterpart of ChatGPT (Dir+Inta) is 66.4 tokens.
Figure~\ref{fig:cases} also shows an example English document with the corresponding summaries generated by ChatGPT and GPT-4 via different prompts.
Therefore, the conversational LLMs prefer to generate lengthy summaries probably due to the RLHF training process, and the interactive prompt further helps them balance informativeness and conciseness, and significantly improves their zero-shot CLS ability.

\vspace{0.5ex}
\noindent \textbf{Best Zero-Shot LLM vs. Fine-Tuned mBART.}
GPT-4 achieves state-of-the-art zero-shot CLS performance among all LLMs, justifying its superiority. But the model is still slightly worse than the fine-tuned mBART-50 in terms of automatic evaluation metrics.
One possible reason is that zero-shot LLMs are not aware of the text style of the golden summaries when performing zero-shot CLS on each dataset. However, lower automatic scores do not indicate worse performance.
For example, as discussed by~\citet{goyal2022news}, the news summaries generated by GPT-3 achieve lower ROUGE scores than fine-tuned models but higher in human evaluation.
Thus, the comparison between LLMs and fine-tuned mBART-50 in CLS needs human evaluation, which we reserve for the future.

\begin{table}[t]
\centering
\setlength{\belowcaptionskip}{-10pt}
\resizebox{0.48\textwidth}{!}
{
\begin{tabular}{lccccc}
\toprule[1pt]
\multicolumn{1}{c}{\multirow{2}{*}{Method}} & CrossSum      & \multicolumn{2}{c}{WikiLingua} & \multicolumn{2}{c}{XSAMSum} \\
\cmidrule(r){2-2}\cmidrule(r){3-4}\cmidrule(r){5-6} \multicolumn{1}{c}{}                       & En$\Rightarrow$Zh         & En$\Rightarrow$Zh          & En$\Rightarrow$De         & En$\Rightarrow$Zh        & En$\Rightarrow$De        \\ \midrule[1pt]
mBART-50 & 32.7 & 46.6 & 75.4 & 22.3 & 27.9 \\ \midrule[1pt]
Davinci-003 (Dir)                           & 83.3          & 78.5           & 149.1         & 61.8         & 62.5         \\
Davinci-003 (TS)     &     82.1     &    76.2       &   148.6      &  53.4     &    65.8    \\
Davinci-003 (ST)    &    44.7     &    49.1   &    91.7   &  43.4      &   52.1     \\
ChatGPT (Dir)                              & 183.7         & 176.6          & 273.5         & 68.6         & 75.3         \\
ChatGPT (Dir+Inta)                    & 66.4 & 50.0  & 80.7 & 28.7& 42.5   \\
ChatGPT (TS)                    & 155.1         & 82.1           & 149.3         & 48.2         & 60.9         \\
ChatGPT (TS+Inta)                & 63.4  & 46.2    & 70.0   & 30.3 & 41.1   \\
ChatGPT (ST)                        & 132.7         & 94.3           & 124.2         & 54.9         & 68.1         \\
ChatGPT (ST+Inta)                & 57.8  & 50.1    & 71.6    & 29.3  & 37.5  \\ 
GPT-4 (Dir)    &  227.1   &  170.5  &  193.1  &  70.4   & 74.4 \\
GPT-4 (Dir+Inta)    &  102.2  &  58.7  &  75.1  &  30.1   &  38.3 \\
GPT-4 (TS)    &  93.9  & 85.6   &  114.7  &  44.1   & 53.8 \\
GPT-4 (TS+Inta)    & 56.5   & 45.4   &  66.5  &  26.3   & 33.8 \\
GPT-4 (ST)    &  106.6  &  87.8  &  109.5  &  43.6   & 53.7 \\
GPT-4 (ST+Inta)    &  62.7  &  48.0  &  65.1  &  26.7   &  33.3 \\ \midrule[1pt]
\multicolumn{1}{c}{\cellcolor{ggreen!30}{\textbf{Golden}}} & \cellcolor{ggreen!30}{36.1} & \cellcolor{ggreen!30}{50.0} & \cellcolor{ggreen!30}{66.8} & \cellcolor{ggreen!30}{23.9} & \cellcolor{ggreen!30}{29.6} \\ \bottomrule[1pt]
\end{tabular}
}
\caption{The average length (token level) of the generated summaries on the test set of each CLS dataset. \colorbox{ggreen!30}{Light green} indicates the length of golden summaries.}
\label{table:output_lens}
\end{table}

\begin{table*}[t]
\centering
\resizebox{0.98\textwidth}{!}
{
\begin{tabular}{lcccccccc}
\toprule[1pt]
                         & \multicolumn{4}{c}{XSAMSum (En$\Rightarrow$Zh)} & \multicolumn{4}{c}{XSAMSum (En$\Rightarrow$De)} \\
                         & Coherence  & Relevance  & Consistency  & Fluency  & Coherence  & Relevance  & Consistency  & Fluency  \\ \midrule[1pt]
\cellcolor{ppink!20}{mBART-50}                 & 54.0      & 32.3       & 36.6         & 55.8     & 54.6      & 36.0      & 45.3         & 52.1     \\ \midrule[1pt]
\cellcolor{bblue!5}{ChatYuan-7.7B (Dir)}      & 44.6       & 17.3       & 40.3         & 53.6     & -          & -          & -            & -        \\
\cellcolor{bblue!5}{ChatYuan-7.7B (Dir+Inta)} & 43.6       & 21.0      & 37.3         & 52.0    & -          & -          & -            & -        \\
\cellcolor{bblue!5}{ChatYuan-7.7B (TS)}       & 41.0      & 17.0      & 24.8         & 48.8     & -          & -          & -            & -        \\
\cellcolor{bblue!5}{ChatYuan-7.7B (TS+Inta)}  & 39.0      & 15.3       & 17.3         & 41.5     & -          & -          & -            & -        \\
\cellcolor{bblue!5}{ChatYuan-7.7B (ST) }      & 47.3       & 19.6       & 36.3         & 54.6     & -          & -          & -            & -        \\
\cellcolor{bblue!5}{ChatYuan-7.7B (ST+Inta)}  & 48.6       & 15.3       & 33.0        & 52.0    & -          & -          & -            & -        \\
\cellcolor{bblue!5}{ChatGLM-6B (Dir) }        & 58.8       & 31.0      & 49.0        & 61.0    & -          & -          & -            & -        \\
\cellcolor{bblue!5}{ChatGLM-6B (Dir+Inta)}    & 60.6       & 35.3       & 55.1         & 60.8     & -          & -          & -            & -        \\
\cellcolor{bblue!5}{ChatGLM-6B (TS)}          & 52.0      & 22.0      & 25.3         & 54.0    & -          & -          & -            & -        \\
\cellcolor{bblue!5}{ChatGLM-6B (TS+Inta)}     & 55.0      & 31.6       & 46.5         & 58.5     & -          & -          & -            & -        \\
\cellcolor{bblue!5}{ChatGLM-6B (ST)}          & 58.6       & 27.0      & 37.3         & 56.5     & -          & -          & -            & -        \\
\cellcolor{bblue!5}{ChatGLM-6B (ST+Inta) }    & 59.3       & 34.0      & 53.1         & 63.8     & -          & -          & -            & -        \\
\cellcolor{bblue!5}{Vicuna-13B (Dir)  }       & 50.3       & 28.0      & 39.6         & 52.8     & 64.3       & 53.6       & 67.1         & 63.8     \\
\cellcolor{bblue!5}{Vicuna-13B (Dir+Inta)}    & 55.5       & 36.0      & 43.0        & 56.8     & 63.8       & 49.0      & 62.0        & 63.6     \\
\cellcolor{bblue!5}{Vicuna-13B (TS)}          & 57.1       & 44.6       & 57.3         & 56.5     & 68.3       & 55.6       & 69.3         & 66.5     \\
\cellcolor{bblue!5}{Vicuna-13B (TS+Inta) }    & 55.1       & 35.6       & 49.3         & 52.1     & 66.6       & 56.3       & 66.3         & 64.0    \\
\cellcolor{bblue!5}{Vicuna-13B (ST) }         & 54.6       & 33.6       & 46.3         & 56.6     & 65.0      & 54.0      & 62.8         & 62.1     \\
\cellcolor{bblue!5}{Vicuna-13B (ST+Inta) }    & 53.6       & 37.0      & 44.3         & 55.1     & 69.5       & 57.3       & 67.6         & 66.3     \\
\cellcolor{bblue!5}{BLOOMZ-176B (Dir)}        & 53.5       & 38.3       & 44.3         & 54.3     & 63.1       & 51.0      & 61.0        & 63.8     \\
\cellcolor{bblue!5}{BLOOMZ-176B (TS)}         & 52.3       & 37.0      & 37.6         & 53.6     & 59.3       & 48.3       & 61.3         & 58.8     \\
\cellcolor{bblue!5}{BLOOMZ-176B (ST) }        & 54.3       & 37.3       & 44.3         & 55.5     & 59.5       & 48.0      & 60.3         & 60.3     \\ \midrule[1pt]
\cellcolor{bblue!10}{Davinci-003 (Dir)}        & 60.0      & 33.0      & 55.0        & 59.3     & 71.3       & 60.6       & 76.6         & 69.5     \\
\cellcolor{bblue!10}{Davinci-003 (TS)}         & 56.3       & 26.6       & 38.6         & 54.3     & 68.5       & 54.3       & 68.6         & 68.3     \\
\cellcolor{bblue!10}{Davinci-003 (ST)}         & 62.8       & 46.6       & 54.0        & 61.5     & 68.6       & 61.6       & 77.0        & 70.1     \\
\cellcolor{bblue!10}{ChatGPT (Dir)}            & 63.1       & 45.3       & \textbf{70.0}        & 65.3     & 74.0      & 64.0      & 82.0        & 71.6     \\
\cellcolor{bblue!10}{ChatGPT (Dir+Inta)}       & 58.0      & 45.0      & 58.0        & 60.6     & 68.3       & 65.3       & 76.0        & 69.1     \\
\cellcolor{bblue!10}{ChatGPT (TS)}             & 63.0      & 49.6       & 59.0        & 62.8     & 71.5       & 62.0      & 77.6         & 71.8     \\
\cellcolor{bblue!10}{ChatGPT (TS+Inta)}        & 64.5       & 49.3       & 61.3         & 60.5     & 70.0      & 59.0      & 77.0        & 69.8     \\
\cellcolor{bblue!10}{ChatGPT (ST)}             & 64.3       & 51.6       & 64.0        & 62.3     & 72.3       & 63.6       & 77.0        & 74.0    \\
\cellcolor{bblue!10}{ChatGPT (ST+Inta)}        & 64.1       & 51.0      & 60.6         & 65.3     & 69.1       & 60.3       & 73.6         & 67.0    \\
\cellcolor{bblue!10}{GPT4 (Dir)}               & 64.0      & 48.6       & 67.6         & \textbf{67.0}    & \textbf{75.3}       & 68.0      & \textbf{83.6}         & \textbf{74.5}     \\
\cellcolor{bblue!10}{GPT4 (Dir+Inta)}          & 62.0      & 50.6       & 57.3         & 63.1     & 70.0      & \textbf{68.6}       & 77.6         & 70.3     \\
\cellcolor{bblue!10}{GPT4 (TS)}                & \textbf{66.0}      & \textbf{55.3}       & 63.3         & 65.6     & 73.6       & \textbf{68.6}       & 79.0        & 72.0    \\
\cellcolor{bblue!10}{GPT4 (TS+Inta)}           & 62.3       & 48.6       & 59.0        & 63.8     & 65.1       & 60.3       & 69.3         & 68.3     \\
\cellcolor{bblue!10}{GPT4 (ST)}                & 63.0      & 52.6       & 64.0        & 62.3     & 72.8       & 67.3       & 80.3         & 72.3     \\
\cellcolor{bblue!10}{GPT4 (ST+Inta)}           & 60.6       & 46.3       & 56.3         & 63.5     & 70.8       & 62.6       & 79.0        & 70.1     \\ \bottomrule[1pt]
\end{tabular}
}
\caption{Evaluation results (judged by ChatGPT) on XSAMSum. \colorbox{ppink!20}{Pink} denotes the fine-tuned baseline. \colorbox{bblue!5}{Light blue} and \colorbox{bblue!10}{blue} denote the zero-shot performance of open-source and non-open-source LLMs, respectively. ``-'' denotes the model cannot be evaluated in the corresponding dataset.}
\label{table:experiments_chatgpt}
\end{table*}

\vspace{0.5ex}
\noindent \textbf{Limited Performance of Open-Source LLMs.}
For open-source LLMs, \emph{i.e.}, BLOOMZ, ChatGLM-6B, Vicuna-13B and ChatYuan-7.7B, they perform poorly on CrossSum and WikiLingua datasets whose documents typically contain more lengthy content than those of XSAMSum.
Although they perform decently on XSAMSum, there is still a large gap compared to GPT-4.
Thus, we conclude that zero-shot CLS is challenging for LLMs to perform due to its composite nature that requires models to perform summarization and translation simultaneously.
In this situation, we suggest future bilingual or multi-lingual LLM research adopt CLS as a testbed to evaluate the LLMs' capabilities to follow composite instructions as well as combine their different ability.

\subsection{LLM-based Evaluation}

It is worth noting that conducting human evaluation on the generated summaries of both LLMs and fine-tuned models is not trivial since human evaluators can easily realize which summaries are generated by LLMs or fine-tuned models. In this manner, the evaluators may have biases during scoring each summary.
To ensure the fairness of human judgment, \citet{stiennon2020learning} only retain the generated summaries whose length belongs to a certain range, and then collect human judgment on these summaries to minimize the potential evaluation bias caused by summary length.
In our scene, the text styles of LLMs and fine-tuned models are quite different, which might also lead to bias.
Thus, the human evaluation of comparing zero-shot LLMs and fine-tuned models on CLS needs more carefully designed.

\begin{figure}[t]
\centerline{\includegraphics[width=0.48\textwidth]{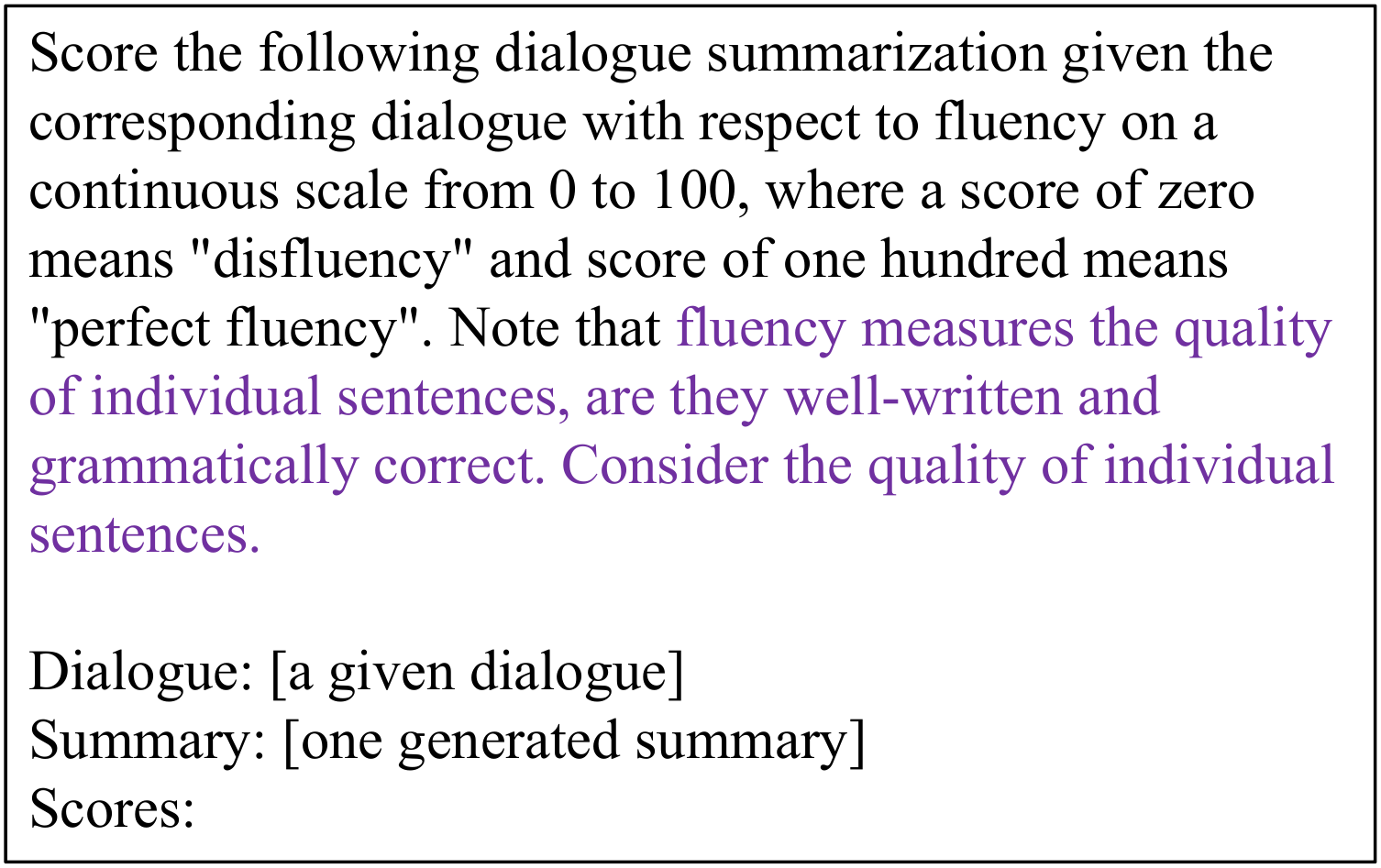}}
\caption{An example prompt used to guide ChatGPT to score the summarization results in the aspect of fluency. \textcolor[RGB]{101,42,150}{Purple} indicates the detailed instruction of the corresponding aspect.}
\label{fig:chatgpt_evaluation_prompt}
\end{figure}

As an alternative to human evaluation, \modi{recent studies~\cite{liu2023gpteval,kocmi2023large,wang2023chatgpt} show that the natural language generation (NLG) results evaluated by LLMs could achieve better correlations with humans.
Following~\citet{wang2023chatgpt}, we utilize ChatGPT to score the generated summaries in a reference-free manner on four aspects, \emph{i.e.}, coherence, relevance, consistency and fluency. An example prompt is shown in Figure~\ref{fig:chatgpt_evaluation_prompt}, and please refer to~\citet{wang2023chatgpt} for prompts of all aspects. The instruction of each aspect (marked in purple in Figure~\ref{fig:chatgpt_evaluation_prompt}) is inspired by SummEval (a widely-used summarization meta-evaluation benchmark dataset)~\cite{10.1162/tacl_a_00373}.
We utilize the official APIs provided by OpenAI\footnote{\url{https://platform.openai.com/docs/guides/gpt/chat-completions-api}} to conduct the experiments with \texttt{gpt-3.5-turbo} model, and set the temperature to 0 to eliminate the randomness of evaluation results.}

\modi{Table~\ref{table:experiments_chatgpt} shows the evaluation results on XSAMSum (En$\Rightarrow$Zh/De). As we can see, GPT-4 achieves the best performance in most aspects, showing its superiority. Besides, compared with the fine-tuned mBART-50 baseline, several zero-shot LLMs, including ChatGLM-6B, Vicuna-13B, Davinci-003, ChatGPT and GPT-4, achieve better results in all aspects, demonstrating the potentiality of performing zero-shot CLS via LLMs. For example, ChatGLM-6B (Dir+Inta) achieves 60.6, 35.3, 55.1 and 60.8 scores in aspects of coherence, relevance, consistency and fluency respectively on XSAMSum (En$\Rightarrow$Zh), while the counterparts of mBART-50 are 54.0, 32.3, 36.6 and 55.8, respectively.
For GPT-4 (Dir), the corresponding scores even reach 64.0, 48.6, 67.6 and 67.0, significantly better than the fine-tuned mBART-50.}

\modi{Moreover, while we show the interactive prompt can improve the performance of zero-shot LLM in terms of ROUGE scores and BERTScore, we do not find the same trend in the LLM-based evaluation results.
In some cases, the interactive prompt even leads to worse LLM-based scores.
We conjecture that the interactive prompt would force zero-shot LLMs to reduce the length of the generated summaries, and the models cannot make a good trade-off between conciseness and other aspects, which is also hard for humans.}

\section{Related Work}

\subsection{Cross-Lingual Summarization}

Given documents in one language, cross-lingual summarization (CLS) generates summaries in another language.
Early work typically focuses on pipeline methods~\cite{Leuski2003CrosslingualCE,Orasan2008EvaluationOA,wan-etal-2010-cross,wan-2011-using,yao-etal-2015-phrase}, \emph{i.e.}, translation and then summarization or summarization and then translation.
Recently, with the availability of large-scale CLS datasets~\cite{zhu-etal-2019-ncls,ladhak-etal-2020-wikilingua,perez-beltrachini-lapata-2021-models,Wang2022ClidSumAB,zheng2022long},
many researchers shift the research attention to end-to-end CLS models.
According to a comprehensive CLS review~\cite{Wang2022ASO}, the end-to-end models involve multi-task learning~\cite{cao-etal-2020-jointly,Bai2021BridgingTG,Liang2022AVH}, knowledge distillation~\cite{Nguyen2021ImprovingNC}, resource-enhanced~\cite{zhu-etal-2020-attend,Jiang2022ClueGraphSumLK} and pre-training~\cite{xu-etal-2020-mixed,chi-etal-2021-mt6} frameworks. However, none of them explore LLMs performance on CLS. To our knowledge, we are the first to explore \emph{can LLMs perform zero-shot CLS} and \emph{how their results are}.

\subsection{Large Language Models}

Recently, there are growing interest in leveraging LLMs for various NLP tasks. \citet{bang2023multitask}, \citet{qin2023chatgpt} and \citet{zhong2023can} conduct systematic investigations of ChatGPT's performance on various downstream tasks. \citet{jiao2023chatgpt} and \citet{peng2023towards} evaluate ChatGPT on machine translation.
\citet{yong2023prompting} show that ChatGPT could generate high-quality code-mixed text.
\citet{tan2023evaluation} explore the performance of ChatGPT on knowledge-based question answering (KBQA).
Some works~\cite{kocmi2023large,wang2023chatgpt,liu2023gpteval,ji2023exploring} utilize ChatGPT or GPT-4 as an evaluation metric to assess task-specific model performance.

\section{Conclusion \modi{and Future Work}}
In this technical report, we evaluate the zero-shot performance of mainstream bilingual and multi-lingual LLMs on cross-lingual summarization.
We find that Davinci-003, ChatGPT and GPT-4 can combine the ability to summarize and translate to perform zero-shot CLS, and achieve competitive results with the fine-tuned baseline (\emph{i.e.}, mBART-50).
In addition, the current open-source LLMs (\emph{i.e.}, BLOOMZ, ChatGLM-6B, Vicuna-13B and ChatYuan) generally show their limited ability to perform CLS in an end-to-end manner, showing the challenge of performing zero-shot CLS still exists.

\modi{In the future, we would like to unleash the potentiality of LLMs and leverage LLMs to perform CLS in few-shot learning manners.}

\section*{Limitations}
While we evaluate the performance of LLMs on the cross-lingual summarization task, there are some limitations worth noting: (1) We only evaluate the lower threshold of these models' CLS performance. Prompts are important to guide LLMs to perform specific tasks, and future work could explore better prompts to obtain better results.
(2) This report only uses two cross-lingual directions (En$\Rightarrow$Zh and En$\Rightarrow$De) in experiments, and all the languages are considered high-resource languages in the world. The performance of LLMs on low-resource languages still needs to be explored. According to~\citet{jiao2023chatgpt}, the machine translation ability of ChatGPT is limited on low-resource languages. We conjecture that the same situation might exist in CLS.
(3) Though the general trend of the evaluation results should be correct, the comparisons between LLMs are not rigorous due to the decoding strategies of these models are not the same.\footnote{Currently, we cannot set the decoding strategy of GPT-4 when manually evaluating it on the ChatGPT platform. Besides, it is difficult to ensure the decoding strategies of LLMs are totally the same when using online platforms or demos.}
This is one of the major reasons leading to the limited soundness of this work.
(4) In the future, we would like to conduct human evaluation to give more analyses.

\section*{Acknowledgement}
We thank anonymous reviewers for their constructive suggestions and comments.

\bibliography{anthology}
\bibliographystyle{acl_natbib}

\appendix

\end{document}